\pdfoutput=1

\documentclass[11pt]{article}

\usepackage{EMNLP2022}

\usepackage{times}
\usepackage{latexsym}

\usepackage[T1]{fontenc}

\usepackage[utf8]{inputenc}

\usepackage{microtype}

\usepackage{inconsolata}

\usepackage{amsmath,amsfonts,amssymb}
\newtheorem{theorem}{Theorem}[subsection]
\newtheorem{corollary}{Corollary}[subsection]
\newtheorem{definition}[theorem]{Definition}

\usepackage{graphicx}
\graphicspath{ {./images/} }
\usepackage{enumitem}
\usepackage{parskip}

\renewcommand{\vec}[1]{\mathbf{#1}}

\usepackage{caption}
\usepackage{subcaption}

\title{Language Model Decomposition: Quantifying the Dependency and Correlation of Language Models}

\author{Hao Zhang \\
  Supportiv Inc, Berkeley, CA \\
  \texttt{haozhang@alumni.princeton.edu} }

\begin{document}
\maketitle
\begin{abstract}
Pre-trained language models (LMs), such as BERT~\citep{devlin2018bert} and its variants, have led to significant improvements on various NLP tasks in past years. However, a theoretical framework for studying their relationships is still missing. In this paper, we fill this gap by investigating the \emph{linear dependency} between pre-trained LMs. The linear dependency of LMs is defined analogously to the linear dependency of vectors. We propose Language Model Decomposition (LMD) to represent a LM using a linear combination of other LMs as basis, and derive the closed-form solution. A goodness-of-fit metric for LMD similar to the coefficient of determination is defined and used to measure the linear dependency of a set of LMs. In experiments, we find that BERT and eleven (11) BERT-like LMs are $91\%$ linearly dependent. This observation suggests that current state-of-the-art (SOTA) LMs are highly ``correlated''. To further advance SOTA we need more diverse and novel LMs that are less dependent on existing LMs.
\end{abstract}

\section{Introduction}
Large-scale pre-trained language models (LMs) have pushed state-of-the-art (SOTA) of NLP recently~\citep{han2021pre, qiu2020pre}. Following BERT~\citep{devlin2018bert}, many variants and improvements have been proposed since then. They differ from BERT in various aspects, in terms of training data~\citep{liu2019roberta}, multilingual support~\citep{conneau2019unsupervised}, model size~\citep{sanh2019distilbert,lan2019albert}, pre-training objective~\citep{yang2019xlnet}, model architecture~\citep{clark2020electra}, attention structure~\citep{he2020deberta}, and sequence length~\citep{beltagy2020longformer}. However, to the best of our knowledge, their relationships have not been studied from a mathematical perspective at the time of writing this paper.

In this work, we present a \emph{quantitative} framework for studying LM \emph{dependency} and \emph{correlation}. Conceptually, we view LM (e.g., encoders like BERT~\citep{devlin2018bert}) as a vector-valued function (or random vector) $\mathbf{u}(x): \Omega \rightarrow \mathbb{R}^d$, where $x$ is text sequence from the sequence space $\Omega$, and $\mathbb{R}^d$ is the sequence embedding space. Sequence embedding is defined as the mean pooling of the last layer's token embeddings. 
We define linear combination and dependency of LMs analogously to vectors/functions, and based upon these definitions we propose the Language Model Decomposition (LMD) algorithm to analyze the linear dependency of LMs.
A goodness-of-fit metric is defined for LMD to quantify the linear dependency and correlation of LMs. In experiments, we find BERT and its successors are highly linearly dependent.

Theoretically, the linear dependency between LMs implies redundancy, since they can be represented by each other. Practically, it simply means building more redundant LMs do not bring in new knowledge. If LMs are more linearly independent, we can potentially distill more knowledge~\citep{alkhulaifi2021knowledge} from multiple diverse models, and combine them to create more powerful models.

The contributions of this paper are: 
(1) We formalize the notion of \emph{linear dependency} for LMs, and propose Language Model Decomposition (LMD) to study linear dependency; 
(2) We define an universal metric based on LMD to quantify the dependency and correlation of LMs.
(3) Our experiments reveal that BERT and its variants are $91\%$ ``correlated'', suggesting current SOTA LMs are highly redundant. The code is available at \url{https://github.com/haozhg/lmd}.

\section{Language Model Decomposition}

\subsection{Notations and Definitions}
In this section, we formalize some necessary mathematical notations and definitions.

\begin{definition}[Linear Combination of LMs]
\label{def:linear-combination}
Given $n$ LMs $\{ \mathbf{u}_i(x) \}_{i=1}^{n}$, a linear combination of these LMs is $\sum_{i=1}^{k} \mathbf{W_i} \mathbf{u_i}(x)$,
where $\mathbf{W_i} \in \mathbb{R}^{d \times d}$ are matrices.
\end{definition}
Notice that here the coefficients are matrices, while for vectors/functions they are scalars.

\begin{definition}[Linear Dependent LMs]
\label{def:linear-dependency}
A set of LMs $\{ \mathbf{u}_i(x) \}_{i=1}^{n}$ is linearly dependent, if there exists matrices $\{ \mathbf{W}_i \}_{i=1}^{n}$ not all singular, such that $\sum_{i=1}^{n} \mathbf{W_i} \mathbf{u_i}(x) = \mathbf{0}$,
$\forall x \in \Omega$, where $\mathbf{0}$ denotes the zero vector.
\end{definition}
 Recall that a matrix $\mathbf{A}$ is singular $\iff$ $\mathbf{A}$ is not invertible $\iff$ $\operatorname {det} (\mathbf{A}) = 0$. If the LMs are not linearly dependent, they are said to be linearly independent, that is, the above equation can only be satisified by singular $\mathbf{W}_i, \forall i$.
 
\begin{corollary}[Linear Dependency Condition]
\label{corollary:linear-dependency-condition}
A set of LMs is linearly dependent if and only if one of them is zero or a linear combination (as in Definition~\ref{def:linear-dependency}) of the others.
\end{corollary}

\subsection{Language Model Decomposition}
To quantify the degree of ``linear dependency'' of a set of LMs, we propose Language Model Decomposition (LMD), where a LM is approximated by a linear combination of other LMs. LMD is motivated by the Galerkin projection method~\citep{reddy2010introduction} that is widely used for model reduction and reduced-order modeling. In particular, given a \emph{target} LM $\mathbf{u}(x)$, and $k$ \emph{basis} LMs $\{\mathbf{v_i}(x)\}_{i=1}^{k}$, we fit a model in the following form
\setlength{\belowdisplayskip}{5pt} 
\setlength{\abovedisplayskip}{5pt}
\begin{equation}
\label{eq:LMD}
    \vec{u}(x) = \sum_{i=1}^{k} \mathbf{W_i} \mathbf{v_i}(x) + \mathbf{e}(x),
\end{equation}
where $\mathbf{e}(x)$ is the residual term. To simplify the derivation, we treat LMs as random vectors from now on. To minimize the residual, we solve the following optimization problem
\begin{equation}
\label{eq:LMDopt}
    \min_{\{\mathbf{W_i}\}_{i=1}^k} L(\mathbf{W_i}) = \operatorname{E} \|\mathbf{e}(x)\|^2,
\end{equation}
where $\operatorname {E}[\cdot]$ is the expectation over $x \in \Omega$, and $\|\cdot\|$ is the $L_2$ norm. $L(\mathbf{W_i})$ is convex (not necessarily strictly convex), so global optimum exists (not necessarily unique). In the special case of $d=1$, it reduces to multivariate linear regression.

\paragraph{Closed-form Solution}
For simplicity, let 
\begin{equation*}
\begin{aligned}
\mathbf{W} & = [\mathbf{W_1}, \mathbf{W_2}, \ldots, \mathbf{W_k}] \in \mathbb{R}^{d \times kd}, \\
\mathbf{z} & = [\mathbf{v_1}^\intercal, \mathbf{v_2}^\intercal, \ldots, \mathbf{v_k}^\intercal]^\intercal \in \mathbb{R}^{kd},
\end{aligned}
\end{equation*}
we can rewrite Eq~\eqref{eq:LMDopt} as
\begin{equation*}
\begin{aligned}
L & = \operatorname{E} [\|\mathbf{u} - \mathbf{W} \mathbf{z}\|^2] \\
              & = \operatorname{E} [ (\mathbf{u} - \mathbf{W} \mathbf{z})^\intercal (\mathbf{u} - \mathbf{W} \mathbf{z})] \\
              & = \operatorname{E} [ \mathbf{z}^\intercal \mathbf{W}^\intercal \mathbf{W} \mathbf{z} - 2 \mathbf{u}^\intercal \mathbf{W} \mathbf{z} + \mathbf{u}^\intercal \mathbf{u} ] \\
            & = \operatorname{E} [ \operatorname{tr} (\mathbf{z} \mathbf{z}^\intercal \mathbf{W}^\intercal \mathbf{W})] - 2 \operatorname{E} [\operatorname{tr} (\mathbf{z} \mathbf{u}^\intercal \mathbf{W})] + c \\
            & = \operatorname{tr} (\operatorname{E} [\mathbf{z} \mathbf{z}^\intercal] \mathbf{W}^\intercal \mathbf{W}) - 2 \operatorname{tr} (\operatorname{E} [\mathbf{z} \mathbf{u}^\intercal] \mathbf{W}) + c,
\end{aligned}
\end{equation*}
where $c=\operatorname{E} [ \mathbf{u}^\intercal \mathbf{u} ]$ is a constant, and we have used cyclic property of trace, linearity of trace, and linearity of expectation. Using the following matrix calculus identity~\citep{petersen2008matrix},
\begin{equation*}
\label{eq:matrixcalc}
    \frac{\partial \operatorname{tr} (\mathbf{AXBX^\intercal C})}{\partial \mathbf{X}} = \mathbf{CAXB} + \mathbf{A^\intercal C^\intercal X B^\intercal},
\end{equation*}
the gradient is
\begin{equation}
\frac{\partial L}{\partial \mathbf{W}} = 2 (\mathbf{W} \displaystyle \operatorname {E}[\mathbf{z} \mathbf{z}^\intercal] - \displaystyle \operatorname {E} [\mathbf{u} \mathbf{z}^\intercal]).
\end{equation}
Setting gradient to zero, the solution is
\begin{equation}
\label{eq:LMDsol}
    \mathbf{W} = \displaystyle \operatorname {E} [\mathbf{u} \mathbf{z}^\intercal] (\displaystyle \operatorname {E} [\mathbf{z} \mathbf{z}^\intercal])^{-1},
\end{equation}
assuming $\operatorname {E} [\mathbf{z} \mathbf{z}^\intercal]$ is full rank (in this case there is an unique global optimal solution). $\operatorname {E} [\mathbf{z} \mathbf{z}^\intercal]$ is the covariance matrix (for simplicity we assume all LMs are mean-subtracted), which is (symmetric) positive semi-definite. Its eigenvalues are real and non-negative. In practice, expectation is approximated with finite samples.

If $\operatorname {E} [\mathbf{z} \mathbf{z}^\intercal]$ is not full rank, Eq~\eqref{eq:LMDopt} is convex but not strictly convex, and there are \emph{infinitely many optimal solutions}. The \emph{minimum-norm} optimal solution is
\begin{equation}
\label{eq:LMDsol-min-norm}
    \mathbf{W} = \displaystyle \operatorname {E} [\mathbf{u} \mathbf{z}^\intercal] (\displaystyle \operatorname {E} [\mathbf{z} \mathbf{z}^\intercal])^{+},
\end{equation}
where $(\displaystyle \operatorname {E} [\mathbf{z} \mathbf{z}^\intercal])^{+}$ is the Moore–Penrose inverse~\citep{moore1920reciprocal} of $\displaystyle \operatorname {E} [\mathbf{z} \mathbf{z}^\intercal]$. Moore–Penrose inverse exists even for non-invertible matrix or non-square matrix. In the special case of a full rank square matrix, Moore–Penrose inverse reduces the ``standard'' matrix inverse.


\paragraph{Regularization}
A small regularization term $\lambda \|\mathbf{W}\|^2$ can added to the loss function in Eq~\eqref{eq:LMDopt}. Mathematically, it ensures that Eq~\eqref{eq:LMDopt} is a strictly convex hence the global optimum is unique regardless of the rank of $\operatorname {E} [\mathbf{z} \mathbf{z}^\intercal]$. Empirically, this increases the numerical stability and accuracy when computing matrix inverse. With regularization, the unique global optimal solution is
\begin{equation}
    \mathbf{W} = \displaystyle \operatorname {E} [\mathbf{u} \mathbf{z}^\intercal] (\lambda \mathbf{I} + \displaystyle \operatorname {E} [\mathbf{z} \mathbf{z}^\intercal])^{-1},
\end{equation}
where $\lambda \mathbf{I} + \displaystyle \operatorname {E} [\mathbf{z} \mathbf{z}^\intercal]$ is positive definite and always invertible.

\paragraph{Bias Term}
In the above derivation, for simplicity we assume all LMs are mean-subtracted. If we include a bias term in LMD equation~\eqref{eq:LMD}, it becomes
\begin{equation}
    \mathbf{u}(x) = \sum_{i=1}^{k} \mathbf{W_i} \mathbf{v_i}(x) + \mathbf{b} + \mathbf{e}(x),
\end{equation}
where $\mathbf{b} \in \mathbb{R}^{d}$ is the bias term. In this case, the solution is
\begin{equation}
\begin{aligned}
    \mathbf{W} & = \text{cov}(\mathbf{u}, \mathbf{z}) (\text{cov}(\mathbf{z}, \mathbf{z}))^{+}, \\
    \mathbf{b} & = \displaystyle \operatorname {E} [\mathbf{u}] - \mathbf{W} \operatorname {E} [\mathbf{z}],
\end{aligned}
\end{equation}
where $\text{cov}(\mathbf{z}, \mathbf{z}) = \displaystyle \operatorname {E} [\mathbf{z} \mathbf{z}^\intercal] - \displaystyle \operatorname {E} [\mathbf{z}] \displaystyle \operatorname {E} [\mathbf{z}^\intercal]$
is the covariance matrix of $\mathbf{z}$, and
$\text{cov}(\mathbf{u}, \mathbf{z}) = \displaystyle \operatorname {E} [\mathbf{u} \mathbf{z}^\intercal] - \displaystyle \operatorname {E} [\mathbf{u}] \displaystyle \operatorname {E} [\mathbf{z}^\intercal]$ 
is the cross-covariance matrix of $\mathbf{u}, \mathbf{z}$.

\subsection{Quantitative Measure of Dependency and Correlation}

\paragraph{Dependency Between Multiple Language Models}
To measure the goodness-of-fit for LMD, we define $\text{R}^2$, analogous to the coefficient of determination used in linear regression~\citep{draper1998applied}. In particular, 
\begin{equation}
\label{eq:R2}
\text{R}^2(\mathbf{u}, \{\mathbf{v_i}\}_{i=1}^{k}) = 1 - \frac{\text{SSR}}{\text{SST}},
\end{equation}
where $\text{SSR} = \operatorname{E} [\|\mathbf{e}(x)\|^2]$ is the residual sum of squares, and $\text{SST} = \operatorname{E} [\|\mathbf{u}(x) - \operatorname{E}[\mathbf{u}(x)] \|^2]$ is the total sum of squares. Here we view $\text{R}^2$ as a function of the target LM $\mathbf{u}(x)$ and basis LMs $ \{\mathbf{v_i}(x)\}_{i=1}^{k}$. Note that $\text{R}^2 \in [0, 1]$. $\text{R}^2 = 1$ implies that the target model and the basis models are perfectly linearly dependent (by Corollary~\ref{corollary:linear-dependency-condition}). A larger $\text{R}^2$ indicates that target model is well approximated by basis models, and LMs are more linearly dependent. Therefore, $\text{R}^2$ is a quantitative measure of the target LM's dependency on the basis LMs.

\paragraph{Correlation Between Multiple Language Models}
The degree of correlation among a group of $n$ LMs $\{ \mathbf{u}_i(x) \}_{i=1}^{n}$ is defined as
\begin{equation}
\label{eq:rho-group}
    \rho(\{\mathbf{u}_i\}_{i=1}^{n}) = \frac{1}{n} \sum_{i=1}^n \text{R}^2(\mathbf{u}_i, \{\mathbf{u_{-i}}\}),
\end{equation}
where $\{\mathbf{u_{-i}}\}$ are the remaining $n-1$ LMs after removing the \emph{target} LM $\mathbf{u}_i$, which are used as the \emph{basis} LMs. Note that $\rho(\{\mathbf{u}_i\}_{i=1}^{n}) \in [0, 1]$. A value of 1 implies that LMs are linear dependent (by Corollary~\ref{corollary:linear-dependency-condition})) and furthermore each of them can be represented by the rest.

\paragraph{Correlation Between Two Language Models}
The measure of correlation between two LMs is of particular interest, in this case it simplifies to 
\begin{equation}
\label{eq:rho-pairwise}
    \rho(\mathbf{u}, \mathbf{v}) = \frac{1}{2} (\text{R}^2(\mathbf{u}, \mathbf{v}) + \text{R}^2(\mathbf{v}, \mathbf{u})),
\end{equation}
where $\text{R}^2(\mathbf{u}, \mathbf{v})$ is the shorthand for $\text{R}^2(\mathbf{u}, \{\mathbf{v}\})$ when there is only one basis LM. Notice that $\rho(\mathbf{u}, \mathbf{v}) \in [0, 1]$. By definition, $\rho(\mathbf{u}, \mathbf{v})$ is symmetric, i.e, $\rho(\mathbf{u}, \mathbf{v}) = \rho(\mathbf{v}, \mathbf{u})$. However, $\text{R}^2(\mathbf{u}, \mathbf{v})$ is asymmetric in general. $\rho(\mathbf{u}, \mathbf{u}) = 1$, meaning that a LM is perfectly ``correlated'' with itself.

In the special case of $d=1, n=2$, LMD reduces to simple linear regression with a single feature. Furthermore, both Eq~\eqref{eq:R2} and Eq~\eqref{eq:rho-pairwise} reduce to the ``standard'' coefficient of determination, which is the square of the correlation coefficient.

\section{Experiments and Results}

\subsection{Experiments}
\label{sec:experiments}

\begin{table}[]
\centering
\resizebox{0.485\textwidth}{!} {

\begin{tabular}{|c|c|}
\hline
model name     & huggingface checkpoint name        \\ \hline
XLM-R          & xlm-roberta-base                   \\ \hline
M-BERT         & bert-base-multilingual-cased       \\ \hline
Longformer     & allenai/longformer-base-4096       \\ \hline
DeBERTa        & microsoft/deberta-base             \\ \hline
distil-M-BERT  & distilbert-base-multilingual-cased \\ \hline
RoBERTa        & roberta-base                       \\ \hline
XLNet          & xlnet-base-cased                   \\ \hline
BERT           & bert-base-uncased                  \\ \hline
ELECTRA        & google/electra-base-discriminator  \\ \hline
distil-RoBERTa & distilroberta-base                 \\ \hline
distil-BERT    & distilbert-base-uncased            \\ \hline
ALBERT         & albert-base-v2                     \\ \hline
\end{tabular}

}
\setlength{\belowcaptionskip}{-10pt}
\caption{Model name and huggingface checkpoint name}
\label{tab:checkpoints}
\end{table}

\paragraph{Language Models}
BERT~\citep{devlin2018bert} and many of its successors are pre-trained encoder LMs, which take in text sequence and output sequence level embedding (defined as the mean pooling of the last layer's token embeddings). In this work, we consider twelve (12) LMs, including BERT~\citep{devlin2018bert}, distil-BERT~\citep{sanh2019distilbert}, M-BERT~\citep{devlin2018bert}, distil-M-BERT~\citep{sanh2019distilbert}, RoBERTa~\citep{liu2019roberta}, distil-RoBERTa~\citep{sanh2019distilbert}, XLM-R~\citep{conneau2019unsupervised}, XLNet~\citep{yang2019xlnet}, ALBERT~\citep{lan2019albert}, ELECTRA~\citep{clark2020electra}, DeBERTa~\citep{he2020deberta}, and Longformer~\citep{beltagy2020longformer}. The We apply LMD to examine the linear dependency between these LMs.

\paragraph{Data}
We utilize the English Wikipedia~\citep{devlin2018bert}\footnote{https://huggingface.co/datasets/wikipedia} and BooksCorpus~\citep{zhu2015aligning}\footnote{https://huggingface.co/datasets/bookcorpus}, which are used in pre-training BERT~\citep{devlin2018bert}. The sequence length $T$ ranges from 16 to 512 tokens, as determined by the BERT tokenizer. When fed into other models, all sequences are truncated at $T$ tokens as determined by their respective tokenizers.
We randomly sample 512,000 and 51,200 sequences as the train data and test data respectively. In our experiments, we find that further increasing the size of the data does not affect the results much.
By central limit theorem, with enough samples the estimation of expectations (see Eq~\eqref{eq:LMDsol}) will become very accurate. 

\paragraph{Implementation}
We download the pre-trained checkpoints for the considered LMs from the public huggingface model hub~\citep{wolf2020transformers}, see Table~\ref{tab:checkpoints} for the full list of checkpoint names. We ran experiments using PyTorch with a NVIDIA V100 GPU. To improve numerical stability, a small regularization term is added to ensure the minimal eigenvalue of $\lambda \mathbf{I} + \operatorname {E} [\mathbf{z} \mathbf{z}^\intercal]$ is $10^{-6}$. 
In our experiments, the sequence embedding is the mean pooling of the last layer's token embeddings.

\subsection{Results}
\label{sec:results}
The closed-form optimal solution (Eq~\eqref{eq:LMDsol}) depends on expectations, which are approximated using train data. The evaluation metrics $\text{R}^2, \rho$ are reported on the test data. LMD Results on another dataset are in Appendix~\ref{sec:more-results}.

\paragraph{Pairwise Correlation of Language Models}
\label{sec:results-pairwise}

\begin{figure}[htb]
    \centering
    \includegraphics[width=0.485\textwidth]{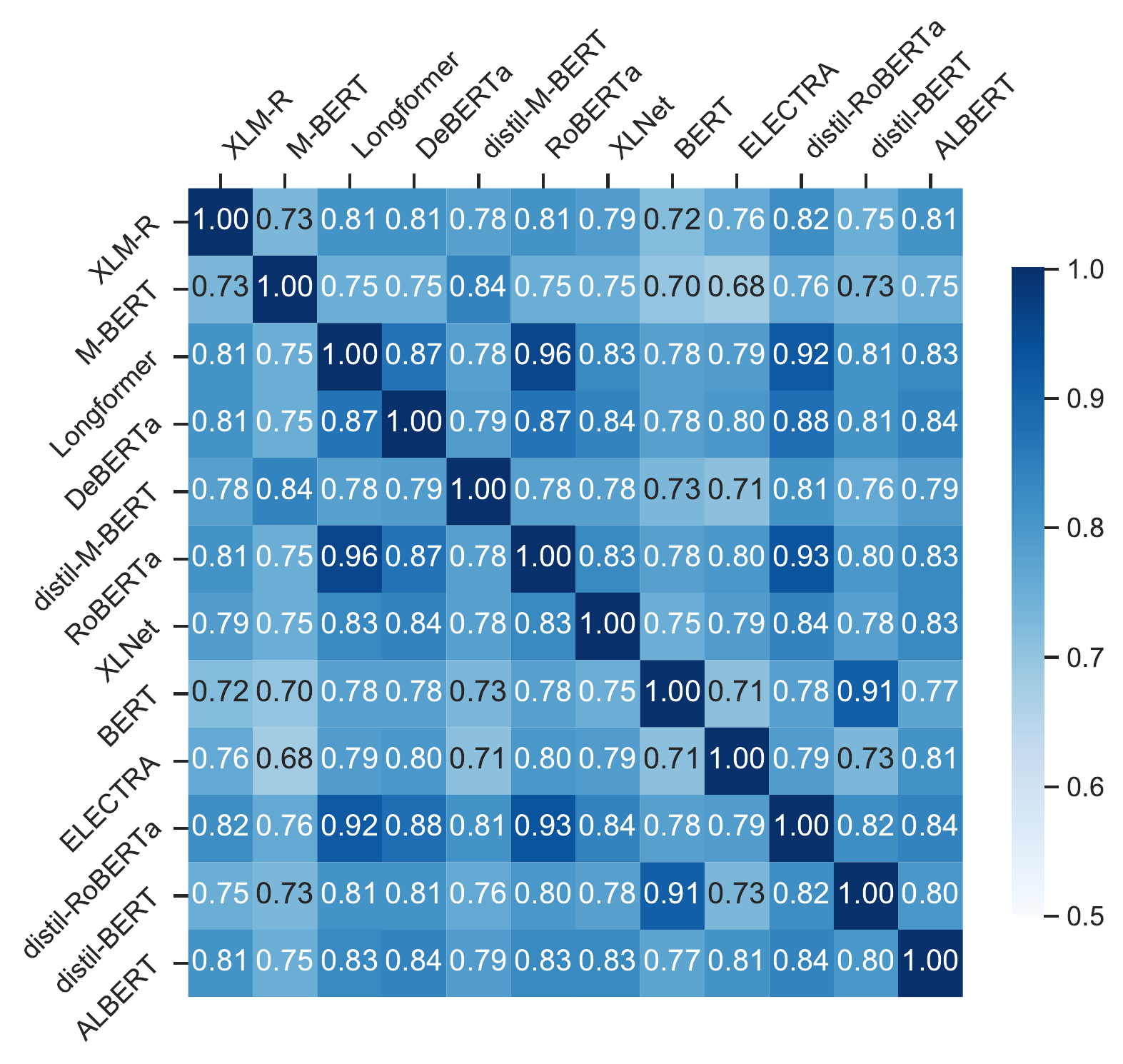}
    \vspace{-20pt}
    \setlength{\belowcaptionskip}{-5pt}
    \caption{Results on \texttt{English Wikipedia} and \texttt{BooksCorpus}. Symmetric pairwise correlation $\rho(\mathbf{u}_i, \mathbf{u}_j)$ (as defined in Eq~\eqref{eq:rho-group}) among 12 language models. The text sequence length is $T=512$.}
    \label{fig:rho-pairwise}
\end{figure}


The symmetric pairwise correlation $\rho(\mathbf{u}_i, \mathbf{u}_j)$ between LMs is visualized in Figure~\ref{fig:rho-pairwise} (for $T=512$), and the asymmetric dependency measure $\text{R}^2(\mathbf{u}_i, \mathbf{u}_j)$ is shown in Figure~\ref{fig:R2-pairwise} of Appendix~\ref{sec:results-details}.
First, notice that distilled models (distil-M-BERT, distil-RoBERTa, and distil-BERT) are highly ``correlated'' with the orignal full models, because they distil knowledge~\citep{alkhulaifi2021knowledge} from the full model.
Second, multilingual LMs (XLM-R and M-BERT) have lower correlation with other LMs. Our hypthesis is that multilingual LMs are trained to generalize to multiple languages, therefore they contain knowledge beyond a single language.

\paragraph{Dependency and Correlation of Multiple Language Models}
\label{sec:results-group}

\begin{figure*}[htb]
    \centering
    \includegraphics[width=1.0\textwidth]{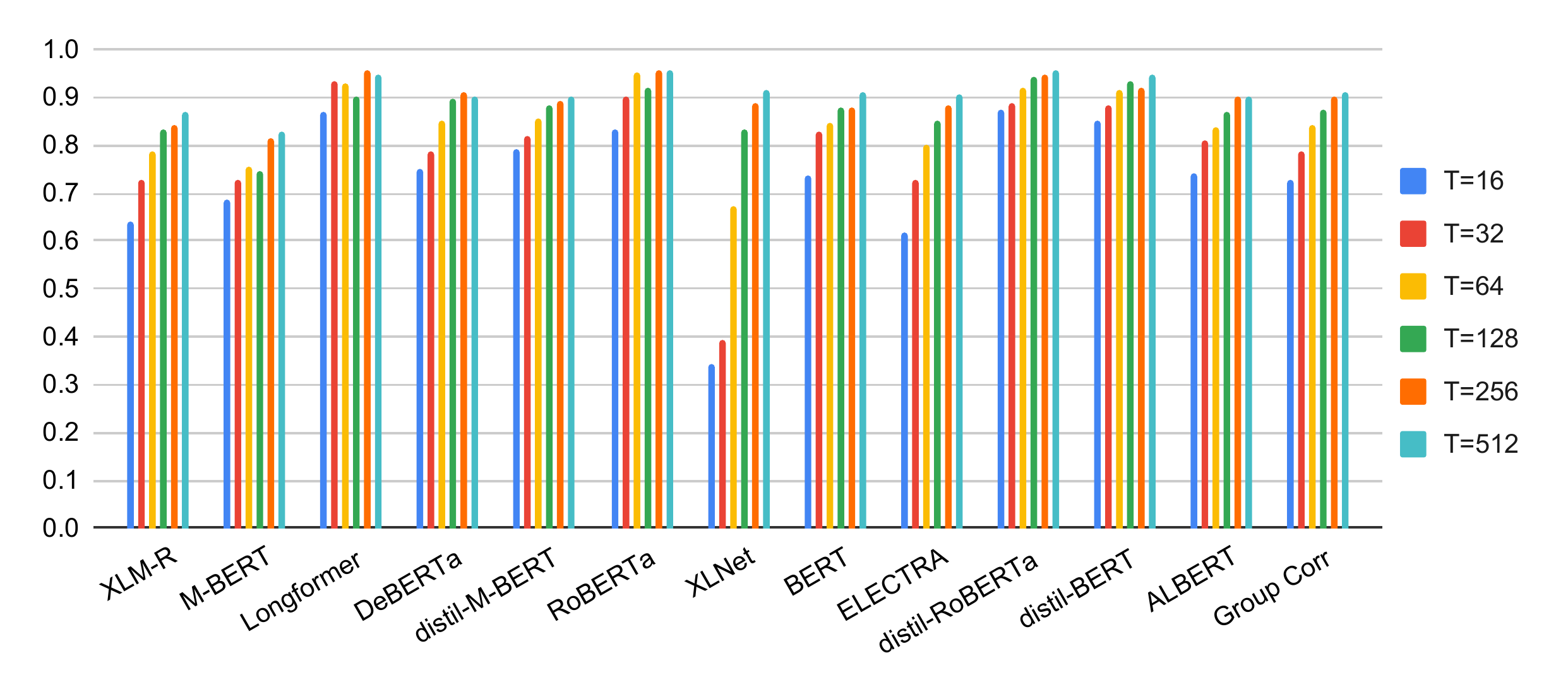}
    \vspace{-20pt}
    \setlength{\belowcaptionskip}{-10pt}
    \caption{Results on \texttt{English Wikipedia} and \texttt{BooksCorpus}. Dependency $\text{R}^2(\mathbf{u}_i, \{\mathbf{u_{-i}}\})$ (as defined in Eq~\eqref{eq:R2}) of language models. The \texttt{Group Corr} refers to the group correlation measure $\rho(\{\mathbf{u}_i\}_{i=1}^{n})$ (as defined in Eq~\eqref{eq:rho-group}). $T$ is the text sequence length. We take each model as \emph{target} model, and apply LMD using the remaining eleven (11) models as \emph{basis} (as defined in Eq~\eqref{eq:LMD}).}
    \label{fig:R2-group}
\end{figure*}

The dependency and correlation of multiple LMs $\text{R}^2(\mathbf{u}_i, \{\mathbf{u_{-i}}\})$ is shown in Figure~\ref{fig:R2-group}. For exact numbers, see Table~\ref{tab:R2-group} in Appendix~\ref{sec:results-details}. 
First, note that $\text{R}^2$ is around $90\%$ for all models (for $T=512$). This is consistent with the fact that all models are variants of BERT. The group correlation $\rho$ is $91.25\%$, suggesting that these LMs are highly linearly dependent, and there is some redundancy in them. 
Second, $\text{R}^2$ is smaller for shorter sequence length $T$. We suspect the reason is that the sequence embedding is the mean pooling of the last layer's token embeddings, so by central limit theory the variations in shorter sequence embedding is larger. Therefore it is harder to approximate the target LM using others as basis LMs. 
Third, similar to pairwise correlation, note that multilingual LMs have lower dependency compared with others.
Finally, for XLNet and ELECTRA, $\text{R}^2$ is lower compared to other LMs (especially for shorter sequences). Our hypothesis is that because they have very different training objectives: Permutation Language Modeling~\citep{yang2019xlnet} for XLNet and Replaced Token Detection~\citep{clark2020electra} for ELECTRA. Therefore they encode text differently from the rest of the models which mostly use Masked Language Modeling~\citep{devlin2018bert}.

\section{Conclusion and Future Work}
In this work, we present a theoretical framework to study the relationships between language models (LMs). 
We formalize the definitions of linear combination and linear dependency of LMs. We further propose the Language Model Decomposition (LMD) algorithm to represent one LM using other LMs as basis. A LMD metric is developed to quantify the linear dependency of LMs.
Our experiments show that BERT and its variants are $91\%$ ``correlated". This suggests that there is redundancy in SOTA pre-trained language models. Preliminary results in this paper demonstrate the potential of LMD as a framework to quantitatively analyze the relationships among LMs. Finally, we leave some open questions to motivate future research.

\setlist{nolistsep}
\begin{enumerate}
   \item Can we leverage the linear dependency or independency of LMs to improve pre-training and/or fine-tuning?
   \item Is there any connection between $\text{R}^2$ and LM performance in the pre-training and fine-tuning stage? What does the ``unexplained variance'' (i.e., $1-\text{R}^2$) of each LM represent?
   \item To further improve SOTA, how can we learn complementary language representations that are less linearly dependent on existing LMs?
   \item Are LMs still highly linearly dependent after fine-tuning on downstream tasks? How does their linear dependency change during fine-tuning?
   \item Are multilingual LMs (e.g., M-BERT, XLM-R) linearly dependent on monolingual LMs?
 \end{enumerate}

\section{Limitations}
In our preliminary experiments, we study encoder LMs similar to BERT. It is worthwhile to
investigate other types of pre-trained LMs, including encoder-decoder LMs (e.g, T5~\citep{raffel2020exploring}, BART~\citep{lewis2020bart}),
decoder LMs (e.g, GPT-2~\citep{radford2019language}), pre-BERT models (e.g, ELMo~\citep{peters-etal-2018-deep}, ULMFit~\citep{howard-ruder-2018-universal}), and even domain-specific
LMs (e.g, FinBERT~\citep{araci2019finbert}). The proposed “dependency” and “correlation” measures quantify the
“similarity” between LMs, but there are still open questions related to interpretation. 
Wikipedia and BooksCorpus are used in our experiments, while the results on a "neutral dataset" (such as a large corpus that is
not used in the pre-training of any of the LMs) are also worth examination. We have chosen the mean pooling of the last layers embedding as the text sequence embedding because it is commonly used for
many downstream tasks. This choice makes the LMD algorithm model
agnostic, meaning that it treats LM as a black box. It only requires the final sequence
embedding, but not the intermediate representations. However, for other task,
the token level embedding is very important (e.g, determining the start and end location of the answer for question-answering).
Therefore, layer-wise and token-level (including the [CLS] and [MASK] token) “correlation” is of interest as
well, and further research is needed.

\bibliography{anthology,custom}
\bibliographystyle{acl_natbib}

\section*{Acknowledgments}
The author is grateful for the tremendous support from his wife and Corgi.

\appendix

\section{Details for Experiment Results}
\label{sec:results-details}
In this section, we show more details for experiment results in Section~\ref{sec:results}.

\begin{figure}[h]
    \centering
    \includegraphics[width=0.485\textwidth]{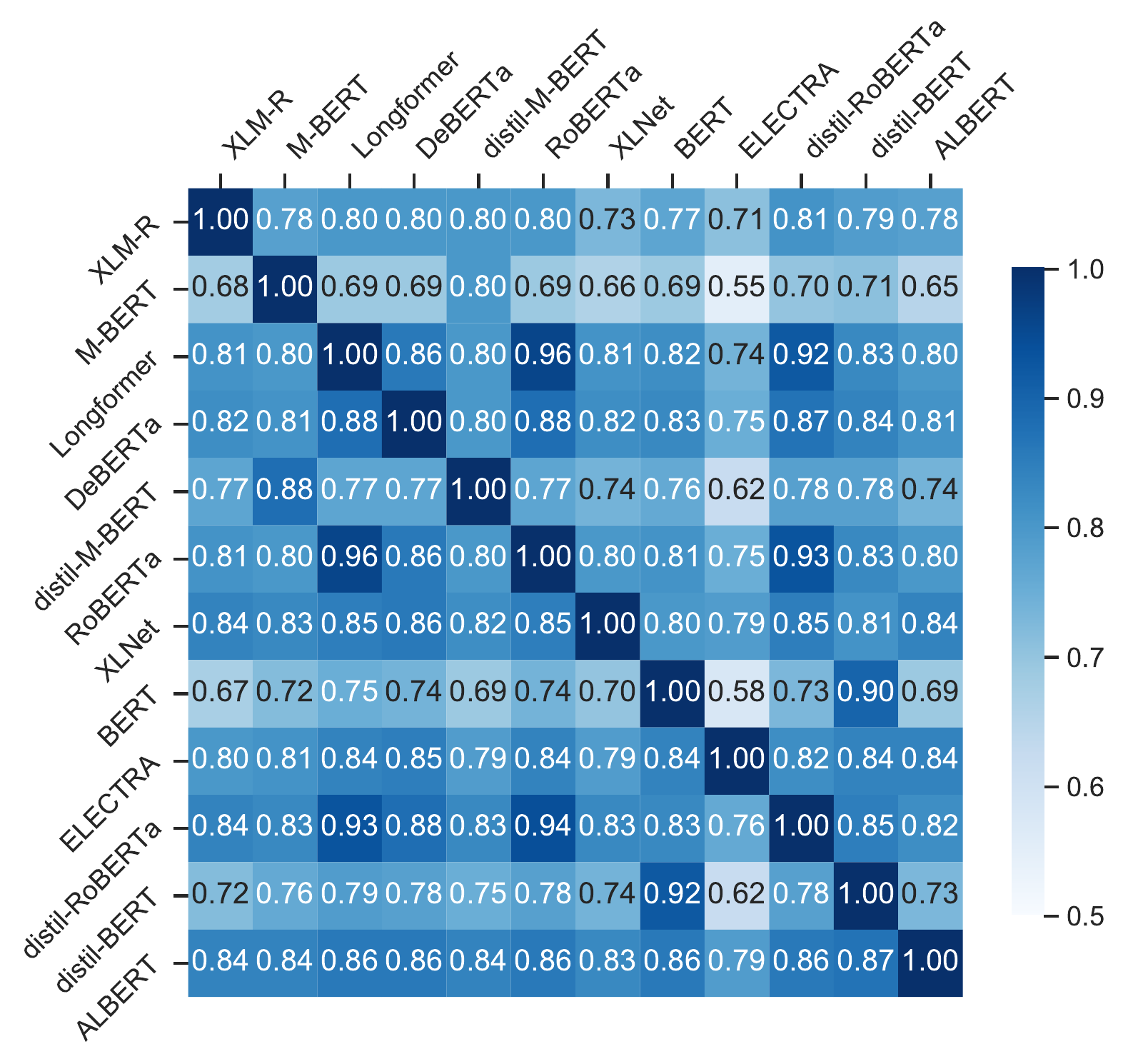}
    \caption{Results on \texttt{English Wikipedia} and \texttt{BooksCorpus}. Asymmetric pairwise dependency $\text{R}^2(\mathbf{u}_i, \mathbf{u}_j)$ (as defined in Eq~\eqref{eq:R2}) of \emph{target} LM (y-axis) on \emph{basis} LM (x-axis). The text sequence length is $T=512$.}
    \label{fig:R2-pairwise}
\end{figure}

\paragraph{Pairwise Dependency of Language Models}
Figure~\ref{fig:R2-pairwise} shows the pairwise asymmetric dependency $\text{R}^2(\mathbf{u}_i, \mathbf{u}_j)$ between language models. Figure~\ref{fig:rho-pairwise} shows the symmetric correlation $\rho(\mathbf{u}_i, \mathbf{u}_j)$.

\paragraph{Dependency and Correlation of Multiple Language Models}
Table~\ref{tab:R2-group} shows the exact numbers for dependency $\text{R}^2(\mathbf{u}_i, \{\mathbf{u_{-i}}\})$ (defined in Eq~\eqref{eq:R2}) between language models. The visualization is shown in Figure~\ref{fig:R2-group}.

\begin{table*}
\centering
\begin{tabular}{|c|c|c|c|c|c|c|}
\hline
               & T=16   & T=32   & T=64   & T=128  & T=256  & T=512  \\ \hline
XLM-R          & 0.6415 & 0.7279 & 0.7866 & 0.8316 & 0.8430 & 0.8692 \\ \hline
M-BERT         & 0.6861 & 0.7278 & 0.7541 & 0.7483 & 0.8139 & 0.8306 \\ \hline
Longformer     & 0.8696 & 0.9329 & 0.9289 & 0.9015 & 0.9581 & 0.9481 \\ \hline
DeBERTa        & 0.7492 & 0.7883 & 0.8503 & 0.8984 & 0.9105 & 0.9038 \\ \hline
distil-M-BERT  & 0.7932 & 0.8190 & 0.8568 & 0.8830 & 0.8940 & 0.9035 \\ \hline
RoBERTa        & 0.8322 & 0.9037 & 0.9512 & 0.9180 & 0.9566 & 0.9552 \\ \hline
XLNet          & 0.3412 & 0.3925 & 0.6733 & 0.8341 & 0.8865 & 0.9154 \\ \hline
BERT           & 0.7381 & 0.8293 & 0.8476 & 0.8803 & 0.8808 & 0.9124 \\ \hline
ELECTRA        & 0.6190 & 0.7278 & 0.8029 & 0.8499 & 0.8852 & 0.9067 \\ \hline
distil-RoBERTa & 0.8722 & 0.8883 & 0.9221 & 0.9417 & 0.9469 & 0.9550 \\ \hline
distil-BERT    & 0.8505 & 0.8827 & 0.9141 & 0.9323 & 0.9199 & 0.9472 \\ \hline
ALBERT         & 0.7422 & 0.8093 & 0.8374 & 0.8696 & 0.9000 & 0.9027 \\ \hline
Group Corr     & 0.7279 & 0.7858 & 0.8438 & 0.8741 & 0.8996 & 0.9125 \\ \hline
\end{tabular}
\caption{
Results on \texttt{English Wikipedia} and \texttt{BooksCorpus}. Dependency $\text{R}^2(\mathbf{u}_i, \{\mathbf{u_{-i}}\})$ (as defined in Eq~\eqref{eq:R2}) of language models. The \texttt{Group Corr} refers to the group correlation measure $\rho(\{\mathbf{u}_i\}_{i=1}^{n})$ as defined in Eq~\eqref{eq:rho-group}. $T$ is the text sequence length. We take each model as \emph{target} model, and apply LMD using the remaining eleven (11) models as \emph{basis} (as defined in Eq~\eqref{eq:LMD}).}
\label{tab:R2-group}
\end{table*}

\section{Additional Experiment Results}
\label{sec:more-results}
We also run LMD using the \texttt{raw English Wikicorpus} dataset~\citep{reese2010wikicorpus}\footnote{https://huggingface.co/datasets/wikicorpus}. This corpus is not directly used in pre-training of aforementioend LMs, though it is in the same domain as the English Wikipedia.

The same group of language models are used (see Section~\ref{sec:experiments}, Table~\ref{tab:checkpoints}), and the train/validation/test sample size is 128,000/12,800/12,800. The regularization parameter is $\lambda=10^{-6}$. We fit the LMD parameters using train set, validate on the validation set, and report dependency and correlation measures on the test set.

\begin{figure*}[h!]
     \centering
     \begin{subfigure}[b]{0.49\textwidth}
         \centering
         \includegraphics[width=\textwidth]{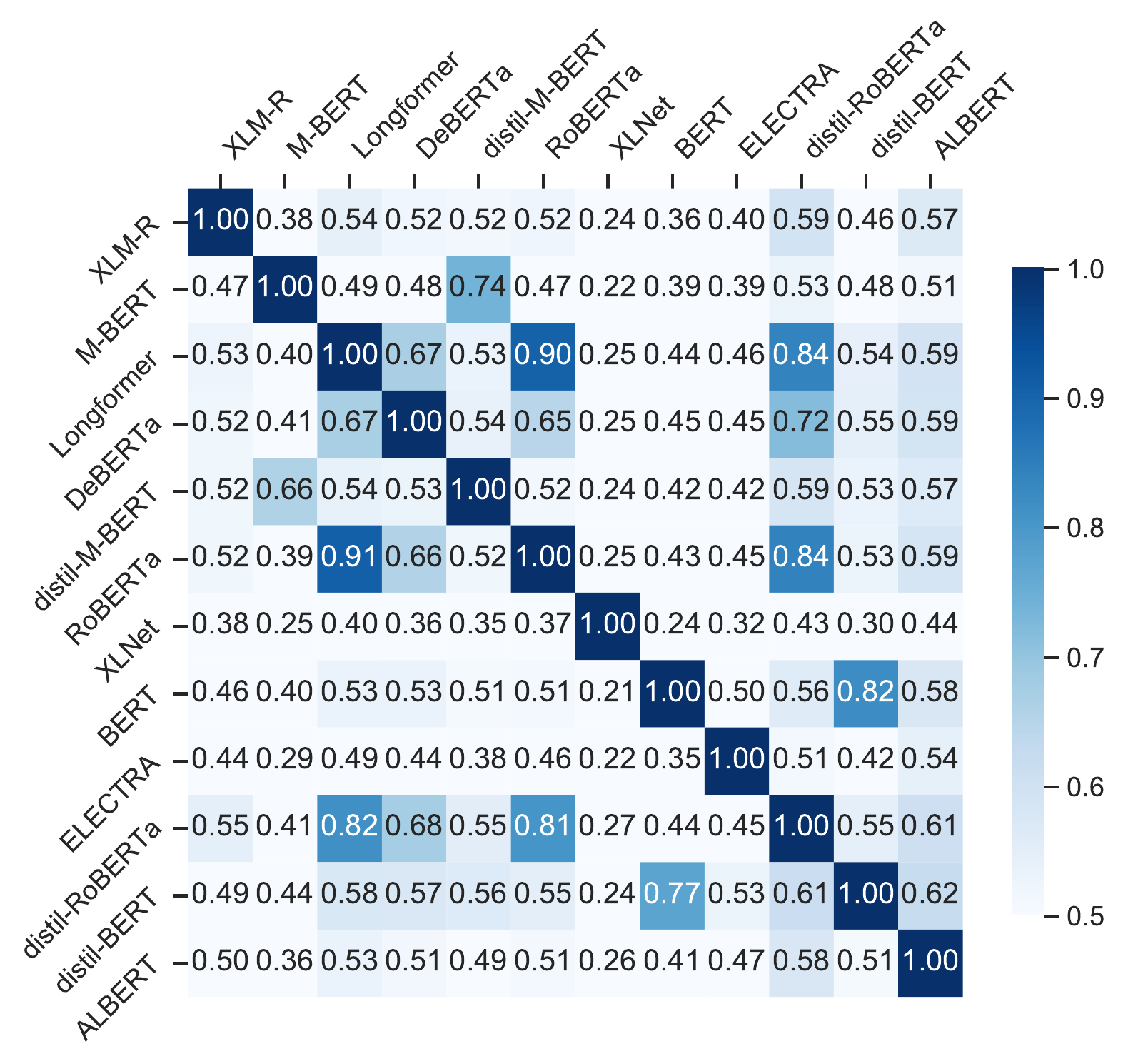}
         \caption{$T=16$}
         \label{fig:wikicorpus-pairwise-R2-16}
     \end{subfigure}
     \hfill
     \begin{subfigure}[b]{0.49\textwidth}
         \centering
         \includegraphics[width=\textwidth]{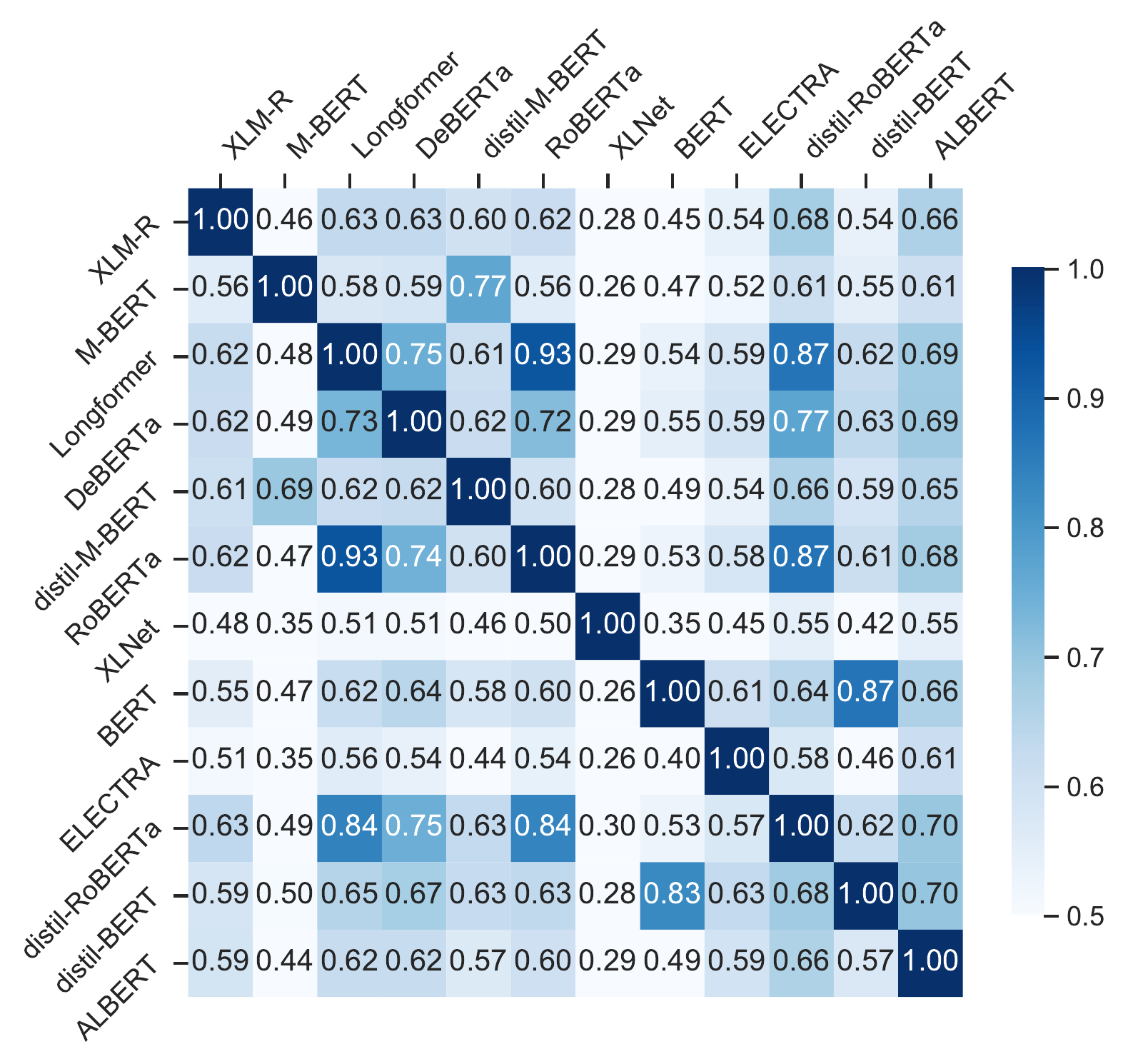}
         \caption{$T=32$}
         \label{fig:wikicorpus-pairwise-R2-32}
     \end{subfigure}
     \hfill
     \begin{subfigure}[b]{0.49\textwidth}
         \centering
         \includegraphics[width=\textwidth]{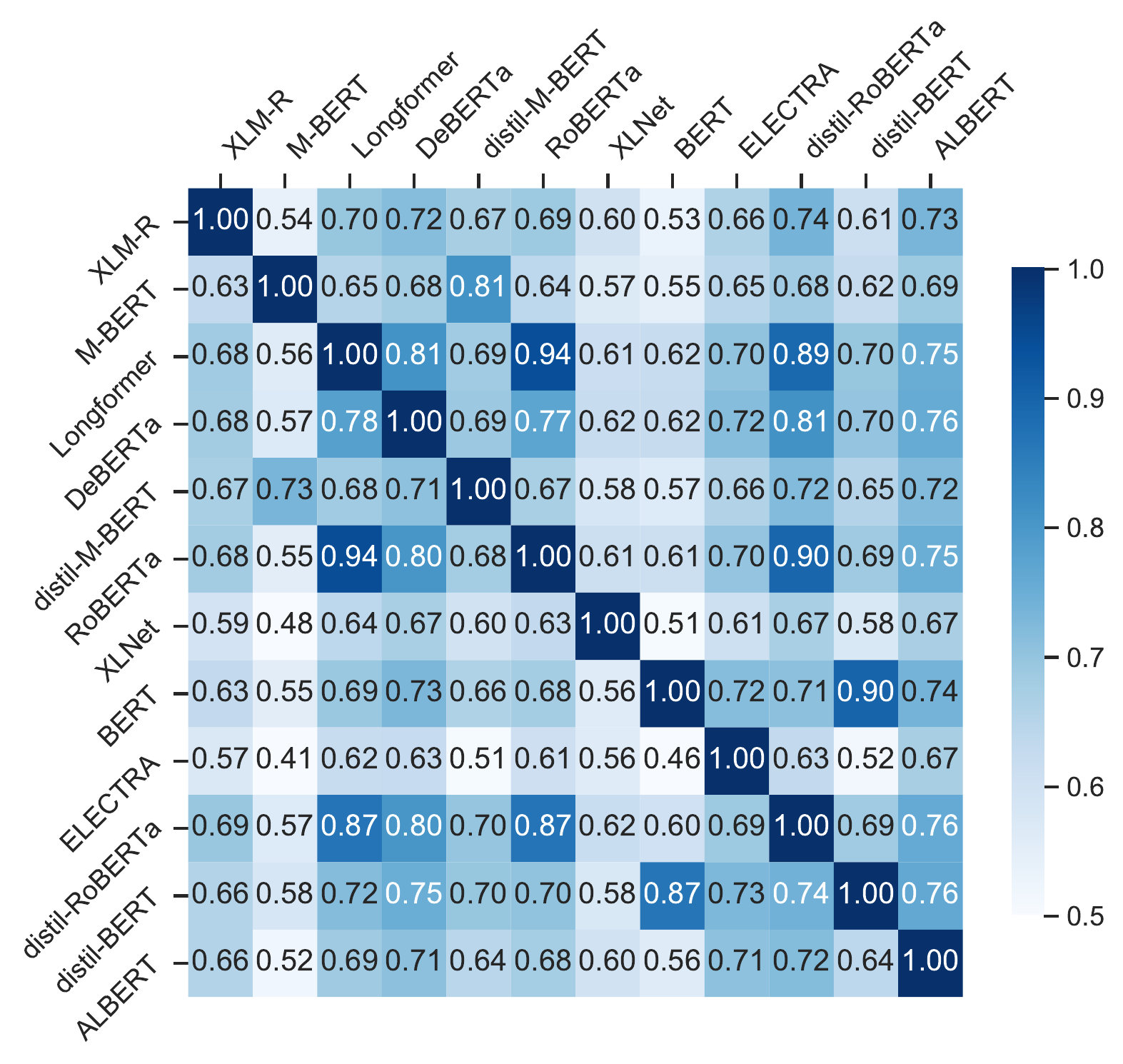}
         \caption{$T=64$}
         \label{fig:wikicorpus-pairwise-R2-64}
     \end{subfigure}
     \hfill
     \begin{subfigure}[b]{0.49\textwidth}
         \centering
         \includegraphics[width=\textwidth]{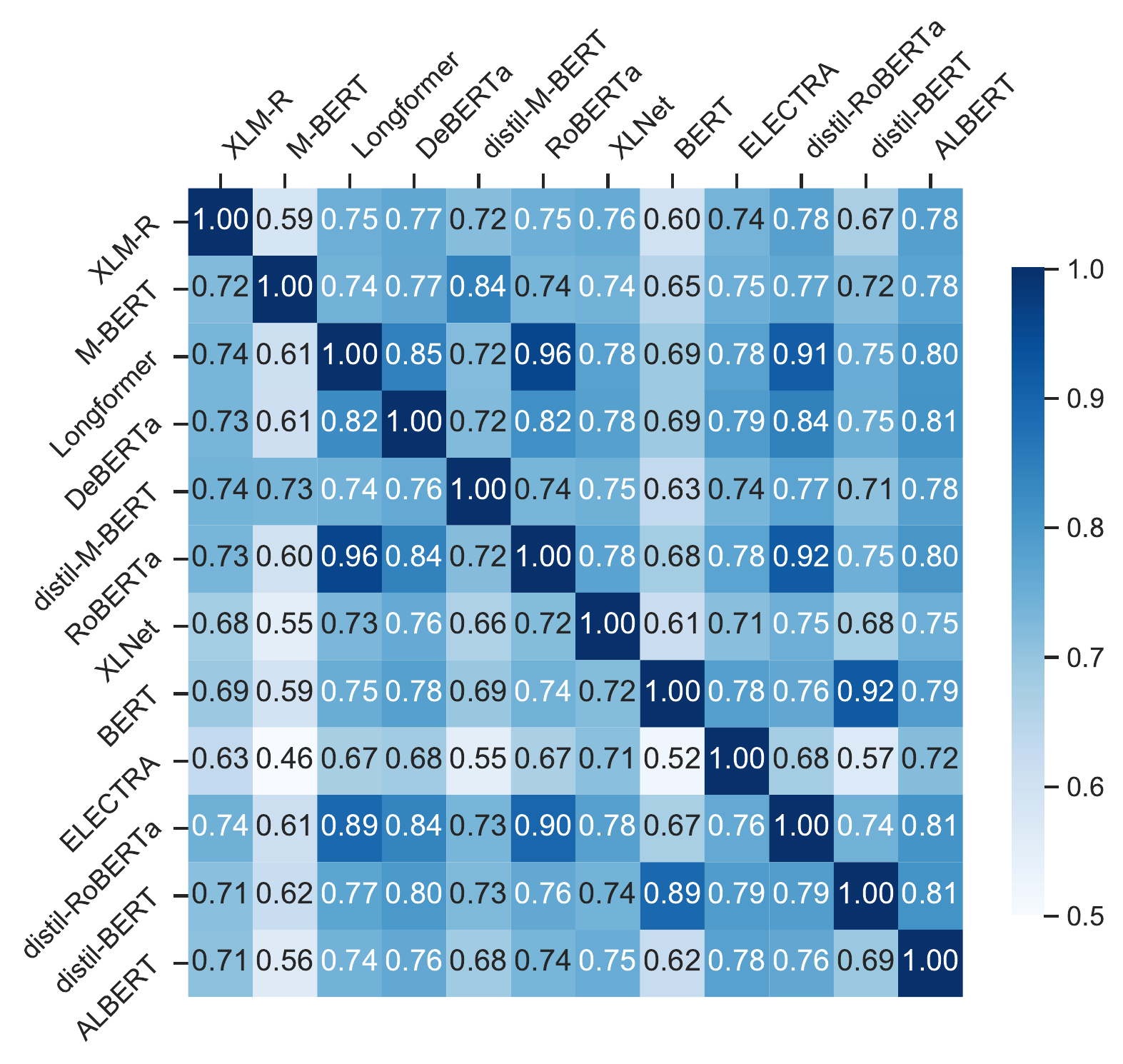}
         \caption{$T=128$}
         \label{fig:wikicorpus-pairwise-R2-128}
     \end{subfigure}
     \hfill
     \begin{subfigure}[b]{0.49\textwidth}
         \centering
         \includegraphics[width=\textwidth]{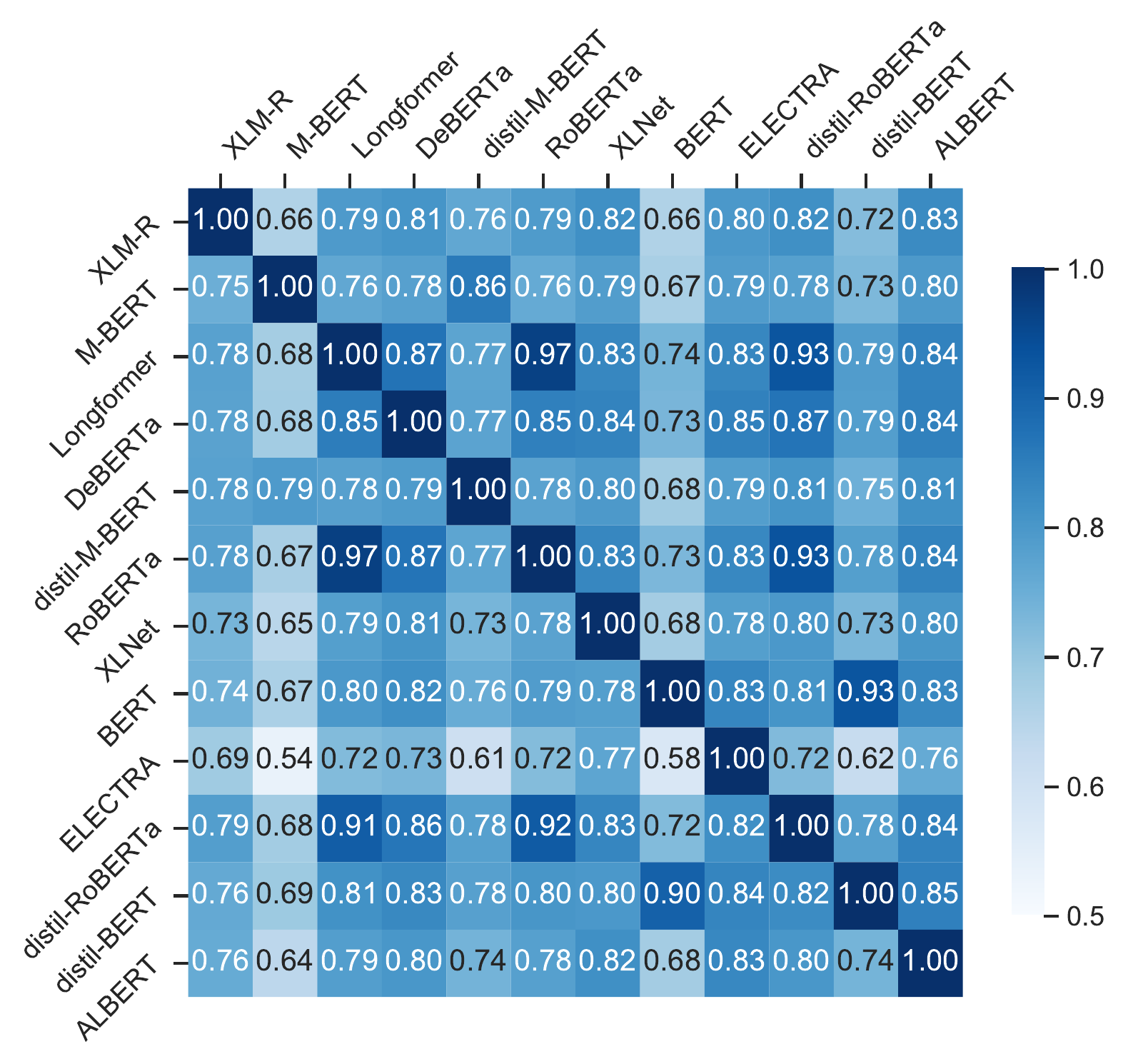}
         \caption{$T=256$}
         \label{fig:wikicorpus-pairwise-R2-256}
     \end{subfigure}
     \hfill
     \begin{subfigure}[b]{0.49\textwidth}
         \centering
         \includegraphics[width=\textwidth]{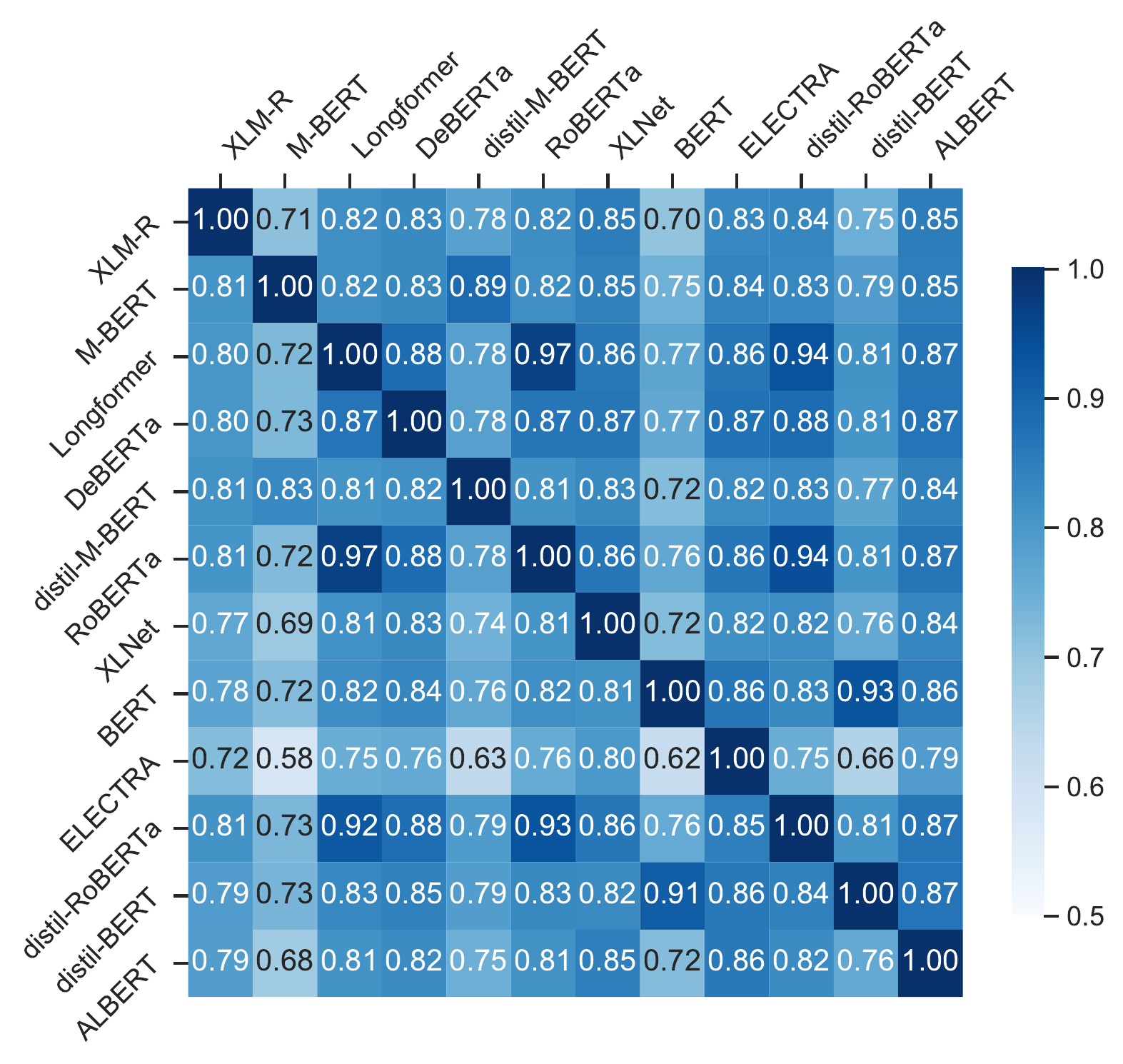}
         \caption{$T=512$}
         \label{fig:wikicorpus-pairwise-R2-512}
     \end{subfigure}
        \caption{Results on \texttt{raw English Wikicorpus}. Asymmetric pairwise dependency $\text{R}^2(\mathbf{u}_i, \mathbf{u}_j)$ (as defined in Eq~\eqref{eq:R2}) of \emph{target} LM (y-axis) on \emph{basis} LM (x-axis).}
        \label{fig:wikicorpus-pairwise-R2}
\end{figure*}

\paragraph{Pairwise Dependency of Language Models}
The pairwise dependency of language models are show in Figure~\ref{fig:wikicorpus-pairwise-R2}. Overall, the dependency $\text{R}^2(\mathbf{u}_i, \mathbf{u}_j)$ (as defined in Eq~\eqref{eq:R2}) increases with sequence length. This provides more evidence for our hypothesis in Section~\ref{sec:results-group}. Because the sequence embedding is the mean pooling of the last layer's token embeddings, by central limit theory the variations in shorter sequence embedding is larger. Therefore it is harder to approximate the target LM using others as basis LMs.

\begin{figure*}
    \centering
    \includegraphics[width=1.0\textwidth]{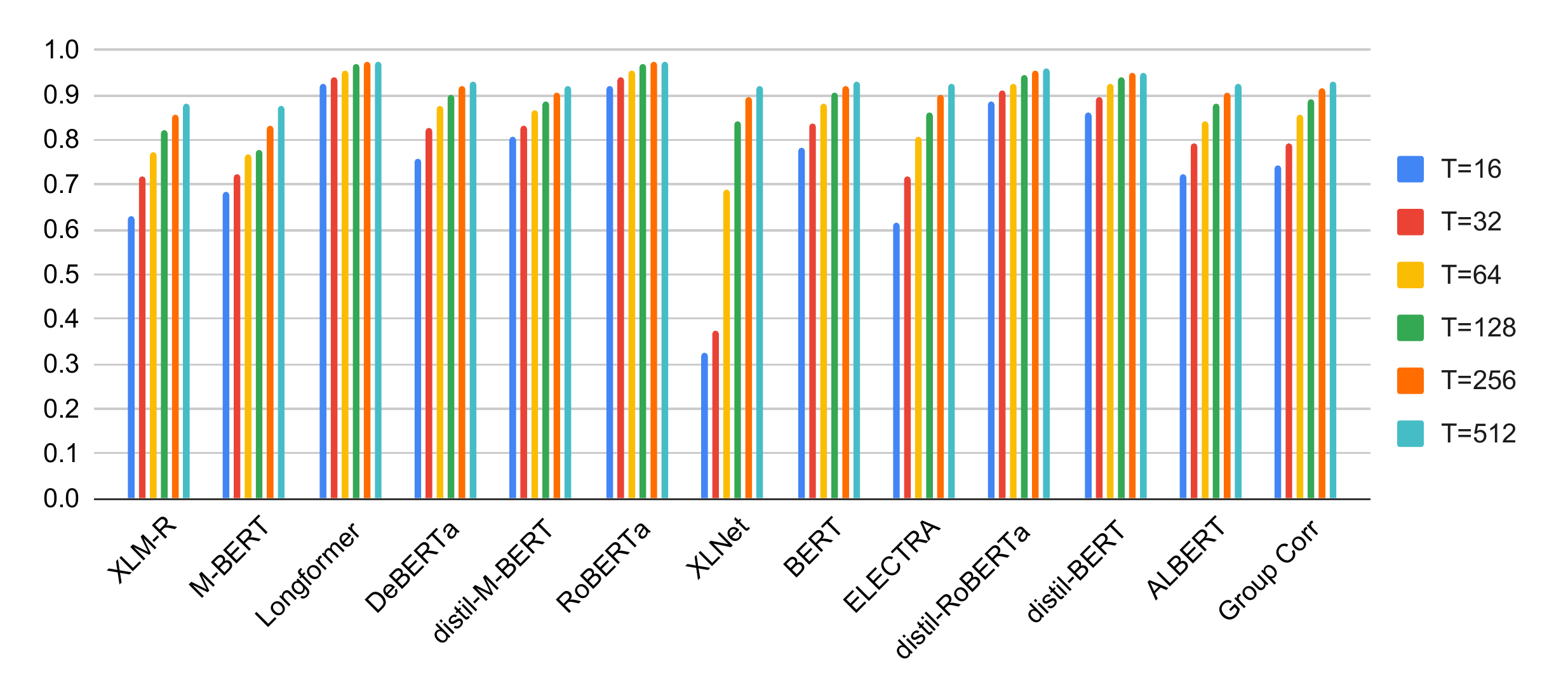}
    \vspace{-20pt}
    \setlength{\belowcaptionskip}{-10pt}
    \caption{Results on \texttt{raw English Wikicorpus}. Dependency $\text{R}^2(\mathbf{u}_i, \{\mathbf{u_{-i}}\})$ (as defined in Eq~\eqref{eq:R2}) of language models. The \texttt{Group Corr} refers to the group correlation measure $\rho(\{\mathbf{u}_i\}_{i=1}^{n})$ (as defined in Eq~\eqref{eq:rho-group}). $T$ is the text sequence length. We take each model as \emph{target} model, and apply LMD using the remaining eleven (11) models as \emph{basis} (as defined in Eq~\eqref{eq:LMD}).}
    \label{fig:wikicorpus-group}
\end{figure*}

\paragraph{Dependency and Correlation of Multiple Language Models}
The dependency $\text{R}^2(\mathbf{u}_i, \{\mathbf{u_{-i}}\})$ (as defined in Eq~\eqref{eq:R2}) and correlation $\rho(\{\mathbf{u}_i\}_{i=1}^{n})$ (as defined in Eq~\eqref{eq:rho-group}) of multiple language models are show in Figure~\ref{fig:wikicorpus-group}, and the exact numbers are in Table~\ref{tab:wikicorpus-group}. The results are similar to the results on \texttt{English Wikipedia} and \texttt{BooksCorpus} (see Figure~\ref{fig:R2-group}, Table~\ref{tab:R2-group}.).

\begin{table*}
\centering
\begin{tabular}{|c|c|c|c|c|c|c|}
\hline
               & T=16   & T=32   & T=64   & T=128  & T=256  & T=512  \\ \hline
XLM-R          & 0.6319 & 0.7179 & 0.7714 & 0.8215 & 0.8562 & 0.8808 \\ \hline
M-BERT         & 0.6847 & 0.7234 & 0.7690 & 0.7753 & 0.8312 & 0.8741 \\ \hline
Longformer     & 0.9240 & 0.9413 & 0.9542 & 0.9668 & 0.9753 & 0.9741 \\ \hline
DeBERTa        & 0.7555 & 0.8260 & 0.8738 & 0.9022 & 0.9202 & 0.9311 \\ \hline
distil-M-BERT  & 0.8048 & 0.8331 & 0.8659 & 0.8843 & 0.9035 & 0.9176 \\ \hline
RoBERTa        & 0.9186 & 0.9401 & 0.9532 & 0.9667 & 0.9750 & 0.9761 \\ \hline
XLNet          & 0.3255 & 0.3766 & 0.6868 & 0.8419 & 0.8959 & 0.9194 \\ \hline
BERT           & 0.7829 & 0.8385 & 0.8784 & 0.9036 & 0.9189 & 0.9283 \\ \hline
ELECTRA        & 0.6143 & 0.7173 & 0.8066 & 0.8618 & 0.8993 & 0.9228 \\ \hline
distil-RoBERTa & 0.8868 & 0.9077 & 0.9264 & 0.9418 & 0.9536 & 0.9612 \\ \hline
distil-BERT    & 0.8593 & 0.8959 & 0.9230 & 0.9396 & 0.9477 & 0.9514 \\ \hline
ALBERT         & 0.7240 & 0.7927 & 0.8435 & 0.8790 & 0.9044 & 0.9223 \\ \hline
Group Corr     & 0.7427 & 0.7925 & 0.8543 & 0.8904 & 0.9151 & 0.9299 \\ \hline
\end{tabular}
\caption{Results on \texttt{raw English Wikicorpus}. Dependency $\text{R}^2(\mathbf{u}_i, \{\mathbf{u_{-i}}\})$ (as defined in Eq~\eqref{eq:R2}) of language models. The \texttt{Group Corr} refers to the group correlation measure $\rho(\{\mathbf{u}_i\}_{i=1}^{n})$ as defined in Eq~\eqref{eq:rho-group}. $T$ is the text sequence length. We take each model as \emph{target} model, and apply LMD using the remaining eleven (11) models as \emph{basis} (as defined in Eq~\eqref{eq:LMD}).}
\label{tab:wikicorpus-group}
\end{table*}

\end{document}